\documentclass[letterpaper]{article}
\usepackage{spconf}

\usepackage{cite}
\usepackage{amsmath,amssymb,amsfonts}
\usepackage{algorithmic}
\usepackage{graphicx}
\usepackage{booktabs}
\usepackage{textcomp}
\usepackage{url}
\usepackage{xcolor}
\let\footnotesize\small

\makeatletter
\newcommand{\ninehalfpt}{%
  \def\baselinestretch{.95}%
  \renewcommand\normalsize{\@setfontsize\normalsize{9.3pt}{11.2pt}}%
  \normalsize
}
\makeatother

\def\BibTeX{{\rm B\kern-.05em{\sc i\kern-.025em b}\kern-.08em
    T\kern-.1667em\lower.7ex\hbox{E}\kern-.125emX}}
\begin{document}

\title{Piano2Pop: Expanding Piano Demos into Pop Productions Without Aligned Supervision}

\name{Yun-Chen Cheng \qquad Chih-Pin Tan \qquad Tzu-Hung Huang \qquad Yi-Hsuan Yang}
\address{National Taiwan University, Taiwan \\
\{hhhhaura, tanchihpin0517, zionhuang1107, affige\}@gmail.com}

\ninehalfpt
\maketitle

\begin{abstract}
We introduce Piano2Pop, a music-expansion task that transforms piano MIDI into instrumental pop productions. 
Enabled by recent advances in music generation and polyphonic music transcription, 
our pseudo-supervised approach conditions a pretrained flow-matching model on piano rolls without
aligned piano--pop supervision. We train on synthetic piano--instrumental pairs derived from pop 
recordings, comparing pair-free transcription, including function-selective (FST) and register--density
(RDA) augmentations, with a pair-supervised piano-cover generator as pseudo-demo sources. On 476 testing
real-world piano-cover/pop pairs, the pair-supervised approach achieves the best Fr\'echet Audio Distance,
Accompaniment Prompt Adherence, and spectral shape on real-piano inputs, while FST mitigates piano-heavy
outputs in pair-free training and remains a promising alternative. All fine-tuned systems
significantly outperform the zero-shot baselines, which suggests a role for task-specific modeling.
\end{abstract}


\section{Introduction}
A pop song often begins as a demo: a guitar sketch, a topline sung over an existing track, or a piano accompaniment. Piano is particularly useful because two hands can convey melody, harmony, voicing, dynamics, and groove, allowing a writer to commit to what the song is without fully specifying what it will sound like. Turning this sketch into a full production requires the producer to infer its musical intent and realize it through drums, instrumentation, arrangement, and mixing.

We call this task \textbf{Piano2Pop}: given a piano MIDI demo, generate a
complete instrumental pop production that is coherent, idiomatic to pop, and
faithful to the demo's musical intent. While Piano2Pop has received little
attention, the reverse problem of mapping pop audio to piano, also called
piano cover generation, has been
explored~\cite{choi2023pop2piano,tan2024picogen2}.
Viewed through instrumentation, those systems perform music
reduction~\cite{chiu2009automatic}, collapsing a multi-instrument mix into
piano; Piano2Pop instead performs \textbf{music expansion}, developing the
demo's structure, melody, harmony, rhythm, and instrumentation rather than
merely accompanying an unchanged piano part.

Existing polyphonic MIDI-to-audio systems focus on synthesis and require aligned MIDI--audio pairs~\cite{hawthorne2022multi}. For the piano-to-pop setting, such supervision is scarce and inherently messy: a piano cover and its corresponding pop 
recording differ in duration, tempo, and often overall structure, a mismatch already noted by previous work~\cite{tan2024picogen2}. We argue that this reliance is no longer necessary. 
Recent multi-instrument transcription models \cite{rouard2026muscriptor} can recover symbolic 
content directly from pop audio, and state-of-the-art music generation models \cite{gong2026ace} can render full productions from various conditions. 
Combining the two, we can pioneer a piano-to-pop system that requires no matched piano-pop pairs.

We propose a pseudo-supervised strategy that derives piano MIDI pseudo demos
from unpaired pop instrumentals to form time-aligned training pairs.
We compare two extraction routes, distinguished by whether the generator
was trained on paired pop recordings and piano covers.
\emph{Pair-free extraction} restricts an audio transcription model's
predictions to piano, yielding a rendition of the full mix without
cover-pop supervision~\cite{rouard2026muscriptor}.
\emph{Pair-supervised extraction} uses a piano arrangement model trained on
cover-pop pairs to generate piano
covers~\cite{tan2024picogen2}. Both provide scalable
supervision but preserve different musical information and may differ
substantially from real piano demos, motivating two research~questions:
\begin{enumerate}
    \item Does Piano2Pop benefit from task-specific adaptation over
    existing conditioned music-generation systems, as measured by
    piano--production adherence and spectral-shape~matchness?
    \item How does pair-free versus pair-supervised pseudo-demo construction
    affect transfer to real piano covers, measured by closeness to real-pop
    audio distributions?
\end{enumerate}

We contribute by formulating Piano2Pop and addressing these research
questions through a piano-conditioned generation framework and an objective
evaluation protocol with tailored behavioral metrics. Audio examples are
available on our demo
page.\footnote{\url{https://hhhhaura.github.io/piano2pop}}
For reproducibility, we plan to release open-source code and checkpoints upon
paper publication.

\section{Related Work}

\label{sec:related}
 
\noindent\textbf{Orchestration in the Symbolic Domain.}
Symbolic piano-to-ensemble orchestration is a close task-level predecessor
to Piano2Pop: both expand a piano sketch into a multi-instrument arrangement.
Early work on piano-to-orchestra expansion explored paired piano and orchestral
scores~\cite{crestel2018database}. The scarcity of such
pairs motivated alternatives based on synthetic reductions and
retrieval~\cite{zhao2024structured}.
These methods output symbolic scores, whereas Piano2Pop generates
instrumental audio directly, combining musical arrangement with its sonic
realization, including timbre and production texture.

\noindent\textbf{Temporally Controlled Music Generation.}
MuseControlLite and JASCO use factorized conditionings, such as melody,
dynamics, rhythm, and related time-aligned controls, to steer
generation~\cite{tsai2025musecontrollite,tal2024joint}.
SingSong and StemGen instead generate an accompaniment or stem that
complements and fits a given condition~\cite{donahue2023singsong,parker2024stemgen}.
ACE-Step supports audio-based cover generation and editing, treating
an audio recording as a sketch of the target
production~\cite{gong2026ace}.
Polyphonic MIDI provides a compact, information-rich alternative, preserving
entangled interactions that cannot
always be cleanly factorized~\cite{lin2023content}.
Piano2Pop chooses this form as its piano sketch and expands it into an
instrumental pop production.

\section{Methodology}
As a first study of Piano2Pop, we focus on fixed-length audio segments
rather than full songs: our training pipeline derives piano pseudo demos
from 30-second pop instrumentals, a horizon that often spans a musical
section (e.g., verse or chorus), then uses these synthetic pairs to
fine-tune a pretrained music generator with a piano-roll encoder.
We describe the pipeline below.

\begin{figure*}[t]
    \centering
    \includegraphics[width=0.861\textwidth]{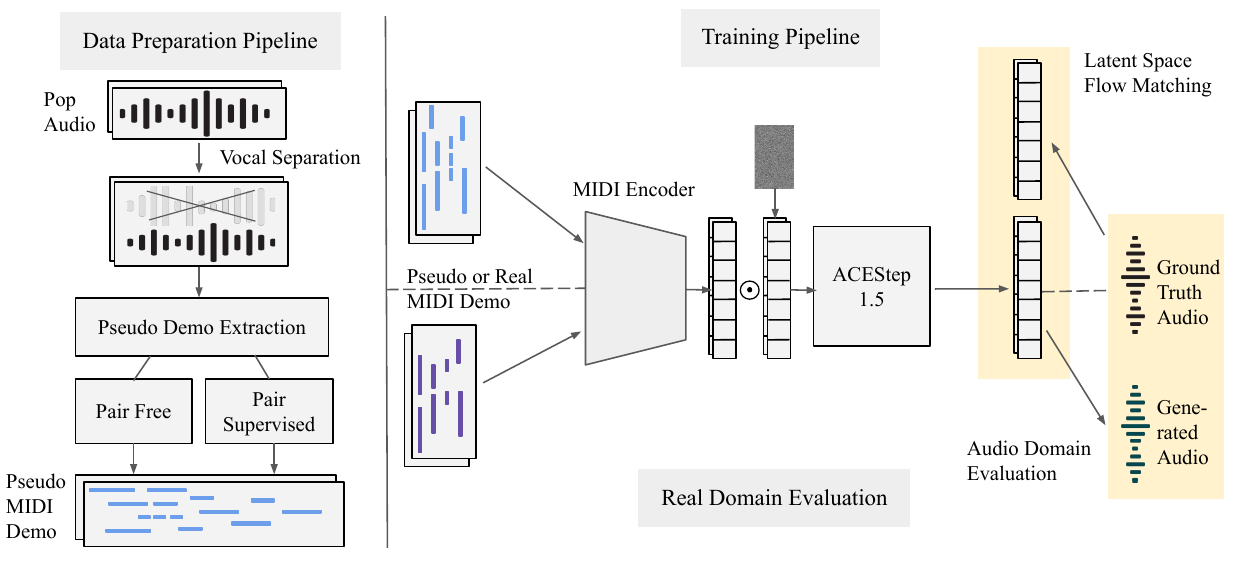}
    \caption{Overview of Piano2Pop.
    \emph{Data preparation:} real pop audio is optionally source-separated,
    transcribed to piano MIDI (pair-free or pair-supervised), and optionally
    augmented.
    \emph{Training:} transcribed MIDI is encoded and fed to a flow-matching
    model that generates audio latents, then decoded to a pop waveform and
    compared with real pop in latent space (transcribed MIDI only).
    \emph{Inference / evaluation:} real piano MIDI follows the same path;
    generated and real pop are compared in the audio domain.}
    \label{fig:pipeline}
\end{figure*}

\subsection{Pair-Free Pseudo-Demo Generation}

We generate pseudo demos with
MuScriptor~\cite{rouard2026muscriptor}, a state-of-the-art multi-instrument
transcription model that outputs multi-track, matrix-like piano rolls
(one track per instrument). Since we only need a piano demo, we force it
to transcribe the full pop instrumental as a piano piece by restricting
decoding to MIDI program~0 (Grand Piano). To assign missing velocities, we
rank notes by onset energy in a $10$~ms framewise log-energy envelope and
map each rank quantile $r_i$ using fixed quantile--velocity anchors
$(\mathbf{q},\mathbf{v})$,
preserving relative dynamics rather than absolute loudness:
\begin{equation}
v_i = \operatorname{clip}\!\left(
\operatorname{interp}(r_i;\mathbf{q},\mathbf{v}),25,95
\right).
\label{eq:rank-velocity}
\end{equation}


To further improve pair-free demos, we observe characteristic relationships
between human piano demos and their paired pop productions, and introduce
two complementary augmentation strategies:

\noindent\textbf{Register--Density Augmentation (RDA).}
Retrospective analysis shows that both pseudo-demo sources have wider local pitch
spreads than real piano, with pair-supervised covers sparser and MuScriptor denser
(Table~\ref{tab:piano-statistics}).
RDA tests whether reducing register and density differences improves
transfer through two stochastic edits: \emph{octave folding} shifts non-bass
notes one octave toward the local register while preserving bassline and
topline boundaries; \emph{function-aware note deletion} thins local harmonic
groups by note function, preferentially retaining structurally important notes.

\noindent\textbf{Function-Selective Transcription (FST).}
Piano arrangement can be viewed as music reduction~\cite{chiu2009automatic}, in which
only selected components of a full arrangement are retained. Following this view,
FST separates each instrumental with HTDemucs~\cite{rouard2023hybrid} into
drums, bass, other, guitar, and piano (vocals excluded), then builds four
mixtures: \textbf{Full} (all stems), \textbf{Harmonic} (pitched stems,
excluding drums), \textbf{No-bass} (all stems except bass), and
\textbf{Piano+bass} (piano and bass only). Each mixture is transcribed as piano.
Before this augmentation, pair-free training uses only a full-mix-to-piano
transcription. After it, Full repeats that demo, while the other three add new
piano-MIDI variants, growing the pair-free training set.

\subsection{Pair-Supervised Pseudo-Demo Generation}
The pair-supervised route uses our previous model,
PiCoGen2~\cite{tan2024picogen2}, trained on paired pop
recordings and piano covers to generate piano arrangements autoregressively
from instrumental features. We generate piano covers of the source
instrumentals, creating pseudo-pairs for the reverse problem: Piano2Pop. Because
PiCoGen2 generates one bar at a time, we beat-track the source instrumental and
linearly stretch each generated MIDI bar to the corresponding beat-derived bar
boundaries. These generated covers provide a consistent, model-derived
arrangement style beyond the local edits provided by our register--density
transformations.

\begin{table}[t]
\centering
\caption{Local pitch spread (30\,ms neighborhoods) and mean note density across 476 matched songs.}
\label{tab:piano-statistics}
\small
\setlength{\tabcolsep}{3pt}
\begin{tabular}{@{}lccc@{}}
\toprule
Statistic & PiCoGen2 & MuScriptor & Real piano \\
\midrule
Pitch spread (semitones) & 19.77 & 19.88 & 16.31 \\
Mean note attacks/s & 3.02 & 7.94 & 5.30 \\
\bottomrule
\end{tabular}
\end{table}

\subsection{Model Architecture}

We fine-tune ACE-Step~v1.5's pretrained rectified-flow transformer
(24 layers, $\approx$2.39B parameters)~\cite{gong2026ace}
with a new piano-roll encoder. A frozen ACE-Step~v1.5 waveform
VAE~\cite{gong2026ace} maps 48\,kHz stereo audio to and from 64-channel,
25\,Hz latents.

\noindent\textbf{Piano-Roll Encoder.}
MIDI is represented as a $176\times T$ tensor at 25\,Hz over pitches 21--108,
with binary sustain and onset-velocity planes (velocities divided by 127).
A five-layer residual 1-D convolutional encoder (kernel 5, width 512,
GroupNorm, SiLU) produces frame-aligned features.
This local receptive field (approximately 0.4\,s per side) focuses 
the encoder on local musical features, leaving global structure modeling
to the diffusion transformer (DiT) backbone.

\noindent\textbf{Conditioning and Fine-Tuning.}
Piano features replace ACE-Step's reference-audio conditioning and are
concatenated with an all-ones generation mask and noisy audio latents
(192 channels total). Encoder and DiT learning rates are $10^{-4}$ and
$10^{-5}$.

\section{Experiments}

\subsection{Training Setup and Zero-Shot Baselines}

We use two disjoint datasets.
\textbf{DB1} is a large collection of real pop songs \emph{without} real piano
MIDI: 8,876 YouTube tracks (8,463 for training, $\approx$554 hours; 413 for
validation), from which we derive pseudo-labeled piano demos for training.
\textbf{DB2} is a smaller set of real pop songs \emph{with} matching real piano MIDI
covers, reserved for evaluation (Section~\ref{sec:eval-data}).

Across all settings, we fine-tune pretrained ACE-Step~v1.5 base for
500,000 steps, training only the piano-roll encoder for the first 2,000 steps
and jointly fine-tuning it with the DiT thereafter.

\noindent\textbf{Pseudo-Demo Construction Ablations.}
We compare pair-free and pair-supervised pseudo-demo generation:
piano-only MuScriptor transcription (\textbf{Base}), function-selective
transcription (\textbf{FST}), MuScriptor transcription with register--density
augmentation (\textbf{RDA}), and PiCoGen2-generated piano covers (\textbf{PiCo}).

We compare the fine-tuned systems with two external baselines using their
released inference pipelines, without our training checkpoints or piano-roll
encoder.

\noindent\textbf{MuseControlLite.}
We use MuseControlLite's released Stable Audio Open
setup~\cite{tsai2025musecontrollite}, conditioned
on monophonic melody, dynamics, and rhythm extracted from the piano audio,
alongside a fixed instrumental-pop text prompt.

\noindent\textbf{ACE-Step cover.}
We use pretrained ACE-Step~v1.5 base's native cover
task~\cite{gong2026ace}, conditioned directly on piano
audio and a fixed pop-arrangement text prompt. We enable instrumental mode,
disable language-model planning, and use 50 inference steps with guidance
scale 7.0.

\subsection{Evaluation Dataset}
\label{sec:eval-data}

\textbf{DB2} provides real piano--pop pairs for evaluation. We construct it by
pairing pop-song piano backing tracks from
Sing2Piano\footnote{\url{https://www.youtube.com/@sing2piano}} and
KaraoKeysPH\footnote{\url{https://www.youtube.com/@KaraoKeysPH}} with their
corresponding instrumentals. Retaining original-key covers with duration
differences of at most $5$~s yields $476$ held-out pairs. These songs are
excluded from DB1 (training and validation), and their piano-cover
channels are disjoint from those used to train PiCoGen2.
For each track and input condition, we evaluate five $30$~s windows, using
strictly matched window starts and deterministic seeds across
systems.

We evaluate two input distributions at inference time. \textbf{Training-distribution inputs}
are synthetic piano MIDI derived from held-out instrumentals using the
corresponding training-data source: MuScriptor transcriptions for Base, FST,
and RDA, and PiCoGen2-generated covers for PiCo. 
\textbf{Real-world inputs} are human piano covers transcribed using the
high-resolution model of Kong et
al.~\cite{kong2021high}, selected for its
ability to estimate note velocities.

\noindent\textbf{Manually Aligned Oracle Subset.}
\label{sec:oracle}
Because the 476 pairs are not temporally verified, we construct a 33-pair
oracle by manually shifting and pitch-preserving time-stretching each
instrumental to its piano. These pairs define the frozen APA reference
$(R,R')$, anchoring a correctly matched piano demo and production.

\begin{table*}[t]
\caption{Five-window objective results (estimate $\pm$ bootstrap SE).
Shape in dB; B+D and Piano in \%. SE is the SD over $B{=}1000$
song-cluster draws. APA uses PCA-$32$ (Section~\ref{sec:apa}).
ACE: ACE-Step cover; MCL: MuseControlLite.
GT APA omitted (476 pairs not temporally aligned); FAD and Shape vs.\ GT
are 0 by construction.}
\label{tab:objective-results}
\vspace*{2pt}
\centering
\footnotesize
\setlength{\tabcolsep}{3.8pt}
\begin{tabular*}{\textwidth}{@{\extracolsep{\fill}}l *{3}{c} *{5}{c}@{}}
\toprule
& \multicolumn{3}{c}{Training distribution} & \multicolumn{5}{c}{Real piano (OOD)} \\
\cmidrule(lr){2-4}\cmidrule(l){5-9}
System & APA$\uparrow$ & FAD$\downarrow$ & Shape$\downarrow$
 & APA$\uparrow$ & FAD$\downarrow$ & Shape$\downarrow$ & B+D (\%) & Piano (\%) \\
\midrule
GT & & &
 & & $.000$ & $.000$
 & $58.58{\pm}1.31$ & $12.22{\pm}1.12$ \\
ACE & & &
 & $.052{\pm}.007$ & $.313{\pm}.002$ & $3.223{\pm}.067$
 & $20.33{\pm}0.57$ & $18.14{\pm}0.57$ \\
MCL & & &
 & $.762{\pm}.006$ & $.647{\pm}.003$ & $3.984{\pm}.070$
 & $3.44{\pm}0.18$ & $53.01{\pm}0.96$ \\
\midrule
Base & $.875{\pm}.010$ & $\mathbf{.263{\pm}.002}$ & $\mathbf{1.534{\pm}.084}$
 & $.857{\pm}.008$ & $.323{\pm}.004$ & $2.756{\pm}.082$
 & $36.09{\pm}0.85$ & $19.69{\pm}0.69$ \\
FST & $\mathbf{.913{\pm}.010}$ & $\mathbf{.263{\pm}.002}$ & $1.906{\pm}.090$
 & $\underline{.895{\pm}.008}$ & $\underline{.255{\pm}.003}$ & $\underline{1.371{\pm}.075}$
 & $51.03{\pm}0.80$ & $12.32{\pm}0.52$ \\
RDA & $.892{\pm}.009$ & $.274{\pm}.002$ & $1.564{\pm}.093$
 & $.849{\pm}.009$ & $.396{\pm}.004$ & $2.587{\pm}.087$
 & $39.78{\pm}0.87$ & $18.28{\pm}0.69$ \\
\midrule
PiCo & $.696{\pm}.009$ & $.279{\pm}.002$ & $2.128{\pm}.076$
 & $\mathbf{.964{\pm}.009}$ & $\mathbf{.231{\pm}.003}$ & $\mathbf{1.044{\pm}.048}$
 & $57.73{\pm}0.73$ & $5.57{\pm}0.32$ \\
\bottomrule
\end{tabular*}
\end{table*}

\subsection{Objective Evaluation Metrics}

To characterize how generated productions compare with real productions, we
measure the following metrics.\footnote{2,380 excerpts (476$\times$5); songs are
independent units. FAD/APA/Shape are recomputed on each of $B{=}1000$
song-cluster draws ($\pm$ bootstrap SE). Zero-shot baselines: real piano only.}

\noindent\textbf{Fr\'echet Audio Distance (FAD).}
FAD compares CLAP embedding distributions of generated and real, unwarped
instrumentals for each input condition.

\noindent\textbf{Accompaniment Prompt Adherence (APA).}
\label{sec:apa}
APA~\cite{grachten2025accompaniment} originally compares
context--accompaniment adherence, defined by aligned references $R$ and
randomly mismatched references $R'$. We adapt it for piano--production
adherence, scoring a candidate distribution $C$ of paired piano demos and
productions against $(R,R')$ with Fr\'echet embedding distance $d$:
\begin{equation}
\mathrm{APA}(C;R,R')=
\frac{1}{2}+
\frac{d(C,R')-d(C,R)}{2d(R,R')},
\label{eq:apa}
\end{equation}
Each pair is encoded as one $512$-d music-only LAION-CLAP embedding of a
single mono mix: the piano window  and the
pop window, summed at matched
loudness. The frozen $(R,R')$ come from the oracle subset
(Section~\ref{sec:oracle}). Because the oracle yields far fewer windows than
the $512$-d embedding size, we fit PCA on $R$ and project $C$, $R$, and
$R'$ onto its top $k{=}32$ components ($\approx$90\% of $R$ variance) before
computing $d$.\footnote{System rankings are unchanged for other $k$
(e.g., $16$ and $64$).}

\noindent\textbf{Level-Normalized Mel-Shape RMSE.}
This diagnostic compares average spectral energy balance independently of overall level. We time-average each excerpt's log-mel spectrum, subtract its
mean across frequency bins, and average over all 2,380 excerpts to obtain
system and unwarped reference profiles:
\begin{equation}
\mathrm{RMSE}_{\mathrm{shape}} =
\sqrt{\frac{1}{K}\sum_{k=1}^{K}
\left(\bar S_{\mathrm{sys}}(k)-\bar S_{\mathrm{ref}}(k)\right)^2},
\end{equation}
where $K=128$ and $\bar S$ is a dataset-average, level-normalized profile in
dB. The metric reflects timbre and mixing/mastering EQ.

\begin{figure}[t]
    \centering
    \includegraphics[width=\columnwidth]{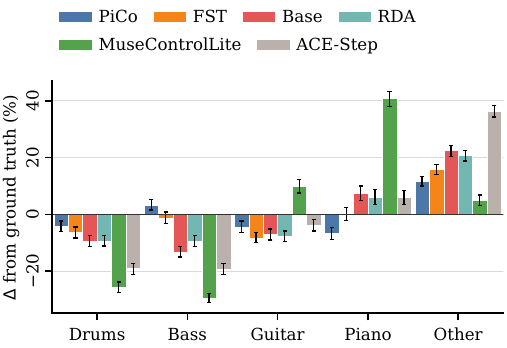}
    \caption{Difference from GT in mean instrument-energy share, with 95\%
    confidence intervals, under OOD real-piano conditioning. Vocals are omitted.}
    \label{fig:instrument-energy}
\end{figure}

\noindent\textbf{Bass+Drums and Piano Energy.}
Piano expansion must reassign notes in the demo: low pitches toward bass,
beats and downbeats toward drums, and the melody toward a lead instrument.
We observed that some systems lack low-frequency energy or stay close to the piano
input. We treat these as failure modes and measure two stem shares.
We separate each excerpt with
HTDemucs~\cite{rouard2023hybrid} and report per-song mean shares of
total power. \textbf{B+D} is the joint bass+drums share; \textbf{Piano} is
the piano-stem share. Both are reported for real-piano inputs only
(Table~\ref{tab:objective-results}); closeness to GT is the target
(Figure~\ref{fig:instrument-energy}).

\section{Experimental Results}
\noindent\textbf{Benefits of Pair-Supervised Pseudo Demos.}
Among fine-tuned systems, PiCo has the best real-piano APA (0.964), FAD
(0.231), and spectral-shape RMSE (1.044; Table~\ref{tab:objective-results}).
Thus, pair-supervised pseudo demos yield the closest adherence,
distributional, and spectral match to real productions under this
evaluation. Its stem profile also indicates stronger transformation
(Figure~\ref{fig:instrument-energy}), having the least difference in
instrument-energy shares from GT.

\noindent\textbf{Augmentations for Pair-Free Pseudo Demos.}
RDA offers no consistent objective gain over Base. FST and Base have
similar training-distribution FAD, but FST is much stronger on real
piano (Table~\ref{tab:objective-results}, Figure~\ref{fig:instrument-energy}), in terms of objective metrics.
In particular, Base and RDA under-produce bass
and drums, whereas FST restores rhythm-section energy closer to GT.
Together, these results suggest that retained musical functions may matter more for transfer than matching
register and density alone.
Though slightly weaker than PiCo on real-piano objectives, FST remains a promising
pair-free alternative with no pop-piano generator.

\noindent\textbf{Baselines versus Fine-Tuned Piano2Pop Models.}
Both zero-shot baselines fail at Piano2Pop. On real-piano inputs, all
fine-tuned systems significantly outperform them on APA and spectral-shape
RMSE ($p{<}0.05$; paired song-cluster bootstrap, 95\% CI excluding zero),
aligning with informal listening. They also significantly improve FAD over
MuseControlLite; PiCo and FST show the best real-piano performance
(Table~\ref{tab:objective-results}, Figure~\ref{fig:instrument-energy}).
ACE-Step cover has near-floor APA and fails to match the input, which might
be due to its audio-latent conditioning: a large stylistic gap from a piano
demo to a full pop production is difficult to cross.
MuseControlLite, however, has the largest real-piano FAD and spectral-shape
error and remains largely piano, despite conditioning on melody, dynamics,
and rhythm, which should better bridge that gap.
We think zero-shot use, not only the stylistic gap, may be limiting and
calls for task-specific modeling.

\section{Conclusion}
We demonstrate Piano2Pop, expanding piano MIDI into instrumental pop
productions through pseudo-supervised fine-tuning of a flow-matching model
on synthetic pairs. On real piano covers, pair-supervised pseudo demos
achieve the best objective results, while function-selective transcription
remains a promising pair-free alternative. Both zero-shot baselines fail,
suggesting that \mbox{Piano2Pop} should be a task of its own and needs
task-specific modeling.

\section{Compliance with Ethical Standards}
This study involved no human participants, personal data, or animals, and no
user study was conducted. We used publicly available YouTube music and
machine-generated audio for research only and do not redistribute copyrighted
audio; institutional ethics approval was not required.

\section{Acknowledgments}
This work is supported by the National Science and Technology Council of
Taiwan (NSTC~114-2628-E-002-013-MY3) and the Ministry of Education (MOE)
of Taiwan (Taiwan Centers of Excellence in Artificial Intelligence). The
authors declare no conflicts of interest. Cursor's GPT-5.6~Sol assisted with
language editing and code; all AI-assisted content was reviewed by the authors.

\bibliographystyle{IEEEbib}
\bibliography{references}

\end{document}